\documentclass[letterpaper]{article} 
\usepackage{natbib}
\setcitestyle{numbers,square}
\usepackage{aaai24}  
\usepackage{times}  
\usepackage{helvet}  
\usepackage{courier}  
\usepackage[hyphens]{url}  
\usepackage{graphicx} 
\usepackage{natbib}  
\usepackage{caption} 
\usepackage{algorithm}
\usepackage{algorithmic}
\newtheorem{myDef}{Definition}
\newtheorem{proposition}{Proposition}
\usepackage{amsmath}
\usepackage{booktabs}
\usepackage{newfloat}
\usepackage{listings}
\DeclareCaptionStyle{ruled}{labelfont=normalfont,labelsep=colon,strut=off} 
\floatstyle{ruled}
\newfloat{listing}{tb}{lst}{}
\floatname{listing}{Listing}
\title{SPD-DDPM: Denoising Diffusion Probabilistic Models in the \\ Symmetric Positive Definite Space}
\author{
    Yunchen Li\textsuperscript{\rm 1}\equalcontrib , Zhou Yu\textsuperscript{\rm 1}\textsuperscript{\rm 2}\equalcontrib, Gaoqi He\textsuperscript{\rm 1},\\
    Yunhang Shen\textsuperscript{\rm 3},
    Ke Li\textsuperscript{\rm 3},
    Xing Sun\textsuperscript{\rm 3},
    Shaohui Lin\textsuperscript{\rm 1}\textsuperscript{\rm 2}\thanks{Corresponding author} ,
}
\affiliations{
    \textsuperscript{\rm 1}East China Normal University\\
    \textsuperscript{\rm 2}Key Laboratory of Advanced Theory and Application in Statistics and Data Science, Ministry of Education, China\\
    \textsuperscript{\rm 3}Youtu Lab, Tencent, Shanghai, China\

    52284404001@stu.ecnu.edu.cn,$\; $ zyu@stat.ecnu.edu.cn,$\; $gqhe@cs.ecnu.edu.cn,\\
    $\{$shenyunhang01,$\; $tristanli.sh,$\; $winfred.sun,$\; $shaohuilin007$\}$@gmail.com
}

\begin{document}

\maketitle

\section{Abstract}

Symmetric positive definite~(SPD) matrices have shown important value and applications in statistics and machine learning, such as FMRI analysis and traffic prediction. Previous works on SPD matrices mostly focus on discriminative models, where predictions are made directly on $E(X|y)$, where $y$ is a vector and $X$ is an SPD matrix. However, these methods are challenging to handle for large-scale data, as they need to access and process the whole data. In this paper, inspired by denoising diffusion probabilistic model~(DDPM), we propose a novel generative model, termed SPD-DDPM, by introducing Gaussian distribution in the SPD space to estimate $E(X|y)$. Moreover, our model is able to estimate $p(X)$ unconditionally and flexibly without giving $y$. 
On the one hand, the model conditionally learns $p(X|y)$ and utilizes the mean of samples to obtain $E(X|y)$ as a prediction.
On the other hand, the model unconditionally learns the probability distribution of the data $p(X)$ and generates samples that conform to this distribution. 
Furthermore, we propose a new SPD net which is much deeper than the previous networks and allows for the inclusion of conditional factors.
Experiment results on toy data and real taxi data demonstrate that our models effectively fit the data distribution both unconditionally and unconditionally and provide accurate predictions.

\section{Introduction}
Symmetric positive definite~(SPD) matrices hold significant importance within the domains of machine learning and multivariate statistical analysis. They play a pervasive role across a diverse range of applications, including FMRI analysis~\cite{petersen2019frechet}, hand gesture recognition ~\cite{nguyen2019neural}, facial recognition~\cite{otberdout2019automatic}, traffic prediction~\cite{tucker2023variable}, brain-computer interfaces~\cite{barachant2013classification}, video classification~\cite{wang2023u,huang2017riemannian,wang2022dreamnet}, interaction recognition~\cite{nguyen2021geomnet}, and many more.

There are two main tasks for SPD matrices: \emph{classification} and \emph{SPD matrix estimation}. The first one involves using SPD matrices as inputs to predict the classes of matrices.
They often involve transforming SPD matrices into Euclidean space vectors via a Log mapping and then employ SPDNet accompanied by convolution with activation~\cite{higham2008functions} and batch normalization~\cite{brooks2019riemannian} for classification. 
%
%
Recently, several works have been proposed to modify SPD Net via multi-scale features, UNet structure, and local convolution~\cite{chen2023riemannian,wang2022dreamnet,zhang2020deep}. 
The second one aims to estimate the SPD matrix as the predicted variable $X$. Previous methods often use traditional statistical modeling frameworks as discriminative models by estimating $E(X|y)$ given the conditional variables $y$. 
For example, they define an additive model in a reproducing kernel Hilbert space using kernel functions~\cite{lin2023additive} or define a regression model in a metric space upon Fréchet mean~\cite{petersen2019frechet,qiu2022random}. However, these models often face challenges when dealing with high-dimensional SPD matrices or predictor $y$. In addition, these methods primarily focus on predicting $E(X|y)$ and currently lack approaches for estimating the probability density in the SPD space. This motivates the construction of generative models for the SPD space. Such models not only benefit from predicting $p(X)$ but also utilize maximum likelihood estimation to make predictions for $X$ given conditional variables $y$.

Denoising diffusion probabilistic model~(DDPM)~\cite{ho2020denoising} is one of the most popular and effective generative models. Several researchers have extended DDPM to Riemannian manifolds, including SO(3)~\cite{leach2022denoising}, SE(3)~\cite{yim2023se}, spheres~\cite{huang2022riemannian}, and more~\cite{de2022riemannian}. These extensions have proven to be of practical value in fields such as protein generation~\cite{corso2022diffdock}, molecular docking~\cite{corso2022diffdock}, and others. However, few works have discussed and studied the SPD space. Meanwhile, the classical DDPM has poor performance when estimating probability distribution in the SPD space. When real data resides on a low-dimensional manifold, estimating in a high-dimensional Euclidean space is inefficient, and the distribution on the manifold may not exhibit favorable characteristics in the Euclidean space where it is embedded.

To address the SPD space distribution estimation problem, we propose a novel denoising diffusion probabilistic model in the SPD space, termed SPD-DDPM. The key components of SPD-DDPM lie in the equation of the backward process and the network to generate SPD matrices. Firstly, by introducing Gaussian distribution, addition and multiplication operations in the SPD space, we extend the DDPM in the Eucidean space to the SPD space, including the forward and backward process. Secondly, we introduce a novel SPD net that takes an SPD matrix as input using double convolutions and allowing for the incorporation of conditions $y$ into the network.

Our contributions are summarized as follows:
\begin{itemize}
	\item We propose a novel SPD-DDPM to accurately estimate the SPD matrix $X$, which provides two versions: conditional and unconditional generation. 
	\item We introduce the Gaussian distribution in the SPD space, the addition and multiplication operations in the SPD space to extend the DDPM theory.
	\item A new SPD U-Net is proposed to effectively incorporate the conditions during the generative process.
\end{itemize}

Experiment results on toy data and real taxi data demonstrate that our models effectively fit the data distribution both unconditionally and unconditionally and provide accurate predictions.

\section{Preliminary}

\subsection{Denoising Diffusion Probabilistic Model~(DDPM)}
Assume the source data $X$ is a distribution $X \sim q(X)$. DDPM defines a forward process of $T$ steps that gradually adds noise to transform the source data into a standard Gaussian distribution by:
\begin{small}
\setlength{\abovedisplayskip}{3pt} 
\setlength{\abovedisplayskip}{3pt} 
    \begin{equation}
		q(X_t|X_{t-1}) = N(\sqrt{1-\beta_t}X_{t-1},\beta_t I),
  		\label{ddpm_eud_forward}
    \end{equation}

\end{small}

%
where $X_t$ is the forward output at the $t$-th step, and $\beta_t$ is a hyperparameter. Eq.~\ref{ddpm_eud_forward} can be reparameterized as:
\begin{small}
\setlength{\abovedisplayskip}{3pt} 
\setlength{\abovedisplayskip}{3pt} 
\begin{equation}
		X_t = \sqrt{1-\beta_t}X_{t-1}+\sqrt{\beta_t}\epsilon_t, \; \epsilon_t \sim N(0,I).
\label{ddpm_eud_reformulate}
\end{equation}
\end{small}

After that, we build the distribution of $q(X_t|X_0)$ as:
%
\begin{small}
\setlength{\abovedisplayskip}{3pt} 
\setlength{\abovedisplayskip}{3pt} 
\begin{equation}
	\begin{split}
		q(X_t|X_0) &= N(\sqrt{\bar{\alpha}_t}X_0, (1-\sqrt{\bar{\alpha}_t})I),\\
		\alpha_t &= 1-\beta_t, \, \bar{\alpha}_t = \prod_{i = 0}^t \alpha_i.
	\end{split}
 \label{ddpm_eud_q(xtx0)}
\end{equation}
\end{small}
Note that the forward process does not require training. After $T$ forward steps, the source data $X$, \emph{a.k.a.} $X_0$, is transformed into a normal distribution. 

In the backward process, DDPM attempts to solve $q(X_{t-1}|X_t,X_0)$ based on Bayes theorem:
\begin{small}
\setlength{\abovedisplayskip}{3pt} 
\setlength{\abovedisplayskip}{3pt} 
\begin{equation}
	\begin{split}
        &q(X_{t-1}|X_t,X_0)  = q(X_t|X_{t-1})\frac{q(X_{t-1}|X_0)}{q(X_t|X_0)} \\
        &\qquad \qquad\qquad\quad= N(\mu_t(X_t,X_0),\sigma^2_tI),\\
        &\mu_t(X_t,X_0)  = \frac{1}{\sqrt{\alpha_t}}(X_t - \frac{1-\alpha_t}{\sqrt{1-\bar{\alpha}_t}}\epsilon), \, \sigma_t^2 = \frac{1-\bar{\alpha}_{t-1}}{1-\bar{\alpha}_t}\beta_t.
    \end{split}
 \label{ddpm_eud_q(xt)}
\end{equation}
\end{small}
where $\epsilon \sim N(0,I)$ represents the noise added to $X_t$, but we do not known its exact value when given $X_t$. Thus, a neural network $\epsilon_{\theta}$ is used to approximate it. The loss function can be formulated as:
\begin{small}
\setlength{\abovedisplayskip}{3pt} 
\setlength{\abovedisplayskip}{3pt} 
\begin{equation}
	\|\epsilon - \epsilon_{\theta}(X_t,t)\|^2
\label{ddpm_eud_object}
\end{equation}
\end{small}	
According to Eq.~\ref{ddpm_eud_q(xt)}, the reverse process is constructed to obtain $p(X_{t-1}|X_t)$:
\begin{small}
\setlength{\abovedisplayskip}{3pt} 
\setlength{\abovedisplayskip}{3pt} 
\begin{equation}
	\begin{split}
		p(X_{t-1}|X_t) &= N\big(\frac{1}{\sqrt{\alpha_t}}X_t-\frac{1-\alpha_t}{\sqrt{\alpha_t(1-\bar{\alpha}_t)}}\epsilon_{\theta},\Sigma_t\big),
	\end{split} \label{ddpm_eud_reverse}
\end{equation}
\end{small}
where $\Sigma_t = \sigma_t^2 I$.

During inference, the data $X_T$ is sampled from the standard Gaussian distribution and then employ Eq.~\ref{ddpm_eud_reverse} to gradually remove noise to generate the desired output of $\hat{X}$.

\subsection{Symmetric Positive Definite~(SPD) Space}
Various real data (\emph{e.g.}, FRMI) are satisfied with the SPD condition to form SPD matrices, which is significantly different from the data in the Euclidean space. In the SPD space (denoted by $Sym^{+}$), symmetry and positive definiteness are held:
\begin{small}
\setlength{\abovedisplayskip}{3pt} 
\setlength{\abovedisplayskip}{3pt} 
\begin{equation}
	\begin{split}
		&Symmetry: \forall X \in Sym^+, X^{'} = X,\\
		&Positive: \forall X \in Sym^+, \forall a \in R^m, a^{'}Xa > 0.
	\end{split}  \label{sym_and_pos}
\end{equation}
\end{small}

Following~\cite{said2017riemannian}, Gaussian distribution in the SPD space is defined as:
\begin{small}
\setlength{\abovedisplayskip}{3pt} 
\setlength{\abovedisplayskip}{3pt} 
\begin{equation}
  p(X|\bar{X},\sigma^2) =\frac{1}{\zeta(\sigma)}\exp[-\frac{d(X,\bar{X})^2}{2\sigma^2}],
\label{gussian_in_spd}
\end{equation}
\end{small}
where $\zeta(\sigma)$ is the regularization coefficient. 
The distance $d$ in Eq.~\ref{gussian_in_spd} is affine-invariant metric~\cite{moakher2005differential}, which is defined as:
\begin{small}
\setlength{\abovedisplayskip}{3pt} 
\setlength{\abovedisplayskip}{3pt} 
\begin{equation}
d(X_1,X_2)^2 = tr[\log(X_1^{-0.5}X_2 X_1^{-0.5})]^2.
\label{affine-invariant}
\end{equation}
\end{small}

Based on the above definition of affine-invariant metric, the exponent and logarithm mappings are formulated by following the work~\cite{higham2008functions}:
\begin{small}
\setlength{\abovedisplayskip}{3pt} 
\setlength{\abovedisplayskip}{3pt} 
\begin{equation}
\begin{split}
		 \operatorname{Exp}_{X_1}(X_2) & = X_1^{\frac{1}{2}}\exp(X_1^{-\frac{1}{2}}X_2X_1^{-\frac{1}{2}})X_1^{\frac{1}{2}}, \\
		\operatorname{Log}_{X_1}(X_2) & = X_1^{\frac{1}{2}}\log(X_1^{-\frac{1}{2}}X_2X_1^{-\frac{1}{2}})X_1^{\frac{1}{2}}.
	\end{split}
 \label{exp_log_mapping}
\end{equation}
\end{small}


In the SPD space, we can employ the exponent and logarithm mappings in Eq.~\ref{exp_log_mapping} to further define the addition and multiplication operations in SPD space for computation:

In the SPD space, the base matrix we select is identity matrix $I$, and the addition operation $\oplus$ and multiplication operation $\odot$ can be defined as:
\begin{small}
\setlength{\abovedisplayskip}{3pt} 
\setlength{\abovedisplayskip}{3pt} 
\begin{myDef}
\begin{equation}
\begin{split}
    X_1 \oplus X_2 &= \operatorname{Exp}_{I}(\operatorname{Log}_{I}(X_1) + \operatorname{Log}_{I}(X_2)).\\
r \odot X & = \operatorname{Exp}_{I}(r \cdot \operatorname{Log}_{I}(X)).    
\end{split}
\end{equation}
\label{plus_mul_in_spd}   
\end{myDef}
\end{small}
Since the base matrix $I$ used in Eq.~\ref{exp_log_mapping}, the exponent and logarithm mappings are equal to $\exp$ and $\log$ function. 
Thus, addition and multiplication operations in \textbf{Definition }~\ref{plus_mul_in_spd} are reformulated as: 
\begin{small}
\setlength{\abovedisplayskip}{3pt} 
\setlength{\abovedisplayskip}{3pt} 
\begin{equation}
	\begin{split}
		X_1 \oplus X_2 &= \exp(\log(X_1)+\log(X_2))\\
		 &= \exp(\frac{1}{2}\log(X_1) +\log(X_2) +\frac{1}{2}\log(X_1))\\
		 & = \exp(\log(X_1^{\frac{1}{2}})+\log(X_2)+\log(X_1^{\frac{1}{2}}))\\
		 &= \exp(\log(X_1^{\frac{1}{2}}X_2 X_1^{\frac{1}{2}})) = X_1^{\frac{1}{2}}X_2X_1^{\frac{1}{2}},
\end{split}\label{spd_plus}
\end{equation}
\end{small}

\begin{small}
\setlength{\abovedisplayskip}{3pt} 
\setlength{\abovedisplayskip}{3pt} 
\begin{equation}
		 r \odot X = \exp(r \cdot \log(X)) = \exp(\log(X^r)) = X^r.
\label{spd_mul}
\end{equation}
\end{small}

In the work~\cite{terras2012harmonic}, SPD space with affine-invariant metric is a homogeneous space under the action of the linear group $GL(m)$, where the group action is defined as:
\begin{small}
\setlength{\abovedisplayskip}{3pt} 
\setlength{\abovedisplayskip}{3pt} 
\begin{equation}
		(X,A) \rightarrow X \cdot A \overset{\underset{def}{}}{=} A^{'}XA.
\label{homogeneous_space}
\end{equation}
\end{small}
Therefore, the affine-invariant metric remains invariant under group actions:
\begin{small}
\begin{equation}
		d(X_1,X_2)^2 = d(X_1\cdot A,X_2\cdot A)^2.
\label{distance}
\end{equation}
\end{small}

Based on Eqs. ~\ref{distance},~\ref{gussian_in_spd}, and ~\ref{affine-invariant}, we have the following proposition:
\begin{small}
\begin{proposition}
    \label{transformation}
    $X \sim G(\bar{X},\sigma^2) \implies X\cdot A \sim G(\bar{X} \cdot A,\sigma^2).$
\label{pro_1}
\end{proposition}
\end{small}
Proof:

For any function $f:P_m \rightarrow R$ ($P_m$ is SPD space), and any $A \in GL(m)$, according to Eq.14, we have:
\begin{equation*}
\small
\setlength{\abovedisplayskip}{3pt} 
\setlength{\abovedisplayskip}{3pt} 
    \int_{P_m} f(X) dv(X) = \int_{P_m} f(X \cdot A)dv(X).
\end{equation*}
where $dv(X)$ is the Riemannian volume element defined as:

\begin{equation*}
    dv(X) = det(X)^{-\frac{m+1}{2}} \prod_{i\le j}dX_{ij}.
\end{equation*}
Intuitively, $dv(X)$ is the Riemannian volume element of $ds^2(X)$, and $ds^2(X)$ is invariant under congruence transformations and inversion:
\begin{equation*}
\small
\setlength{\abovedisplayskip}{3pt} 
\setlength{\abovedisplayskip}{3pt} 
    ds^2(X) = tr[X^{-1}dX]^2.
\end{equation*}

Let $X$ be a random variable in $P_m$, let $\phi:P_m \rightarrow R$ be a test function. If $X\sim G(\bar{X},\sigma^2)$ and $Z=X \cdot A$, then the expection of $\phi(Z) $is given by:
\begin{equation*}
\small
\setlength{\abovedisplayskip}{3pt} 
\setlength{\abovedisplayskip}{3pt} 
\begin{split}
    &\int_{P_m} \phi(X \cdot A)p(X|\bar{X},\sigma^2)dv(X) \\
    &=\int_{P_m}\phi(Z)p(Z \cdot A^{-1}|\bar{X},\sigma^2)dv(Z)\\
    &=  \int_{P_m}\phi(Z)p(Z|\bar{X}\cdot A,\sigma^2)dv(Z).
\end{split}
\end{equation*}


\section{Method}

\subsection{Symmetric Positive Definite Denoising Diffusion Probabilistic Model~(SDP-DDPM)}
DDPM is computed under the Euclidean space, which can not directly be applied to the SPD space.
To this end, we propose SDP-DDPM based on the operational rules (\emph{i.e.}, Eq.~\ref{spd_plus} and Eq.~\ref{spd_mul}) and Proposition 1 in the SPD space.

In the forward process, following DDPM, we can formulate $q(X_t|X_{t-1})$ in the SPD space as: 
\begin{equation}
\small
\setlength{\abovedisplayskip}{3pt}   
\setlength{\belowdisplayskip}{3pt} 
		q(X_t|X_{t-1}) \sim G(\alpha_t \odot X_{t-1},\beta_t^2), 
  \label{ddpm_spd_forward}
\end{equation}
where $\alpha_t^2 +\beta_t^2 = 1,$ and $G(\cdot,\cdot)$ is gaussian distribution in the SPD space defined in Eq.~\ref{gussian_in_spd}. By using Eq.~\ref{spd_plus}, Eq.~\ref{spd_mul}) and  Proposition 1, we have:
\begin{equation}
\small
\setlength{\abovedisplayskip}{3pt}   
\setlength{\belowdisplayskip}{3pt} 
	\begin{split}
		X_t &= \alpha_t \odot X_{t-1} \oplus \beta_t \odot \epsilon_t, \quad\epsilon_t \sim G(I,1)\\
		&= X_{t-1}^{\frac{\alpha_t}{2}} \;\epsilon_t^{\beta_t} \; X_{t-1}^{\frac{\alpha_t}{2}}\\
		& \sim G(\alpha_t \odot X_{t-1},\beta_t^2).
	\end{split}  \label{ddpm_sdp_repara}
\end{equation}
By using Eq.~\ref{ddpm_sdp_repara}, we can obtain the relationship between $X_t$ and $X_0$ as below: 
\begin{equation}
\small
\setlength{\abovedisplayskip}{3pt}   
\setlength{\belowdisplayskip}{3pt} 
	\begin{split}
		X_t &= \alpha_t \odot X_{t-1} \oplus \beta_t \odot \epsilon_t\\
		&= \alpha_t \odot (\alpha_{t-1} \odot X_{t-2} \oplus \beta_{t-1} \odot\epsilon_{t-1})\oplus \beta_t \odot\epsilon_t\\
		&= \alpha_t \alpha_{t-1}\odot X_{t-2} \oplus \alpha_t \beta_{t-1} \odot \epsilon_{t-1} \oplus \beta_t \odot\epsilon_t \\
		&= (\alpha_t\dots \alpha_1)\odot X_0 \oplus (\alpha_t \dots \alpha_2)\beta_1 \epsilon_1 \oplus \dots \oplus \beta_t\epsilon_t\\
		&= \bar{\alpha}_t \odot X_0 \oplus \sqrt{1-\bar{\alpha}_t^2} \odot \epsilon, \; \epsilon \sim G(I,1) \\
		&\sim G(\bar{\alpha}_t \odot X_0,1-\bar{\alpha}_t^2), \; \bar{\alpha}_t = \prod_{i = 0}^t \alpha_i.
	\end{split} \label{x_tx_0}
\end{equation}
The detailed proof of Eq.~\ref{x_tx_0} in the SPD space is provided in the supplementary materials.

By defining $\bar{\beta}_t$ to be $\sqrt{1- \bar{\alpha}_t^2}$, $X_t$ can be further formulated as:
%
\begin{equation}
\small
\setlength{\abovedisplayskip}{3pt}   
\setlength{\belowdisplayskip}{3pt} 
  {X_t = \bar{\alpha}_t \odot X_0 \oplus \bar{\beta}_t \odot \epsilon.}
\label{ddpm_spd_xt}
\end{equation}
Based on Eq.~\ref{ddpm_spd_xt}, we only set $\bar{\alpha}_t$ to approach 0, $X_t$ will be transformed to the standard Gaussian distribution in the SPD space. Empirically, $\alpha_t$ is set to $\sqrt{1-\frac{0.08t}{T}}$ in our experiments, such that $\bar{\alpha}_t \rightarrow{0}$ when $t\rightarrow\infty$. Therefore, we can randomly sample the noise from the standard Gaussian distribution in the SPD space, and use the following backward process to generate the desired output.  


In the backward process, we also compute $q(X_{t-1}|X_t, X_0)$ using the Bayesian technique as:
\begin{equation}
\small
\setlength{\abovedisplayskip}{3pt}   
\setlength{\belowdisplayskip}{3pt} 
	\begin{split}
		&q(X_{t-1}|X_{t},X_0) = q(X_t|X_{t-1})\frac{q(X_{t-1}|X_0)}{q(X_t|X_0)}\\
		& \propto \exp(-\frac{d(X_t,X_{t-1}^{\alpha_{t}})^2}{2 \beta_t^2}-\frac{d(X_{t-1},X_0^{\bar{\alpha}_{t-1}})}{2\bar{\beta}_{t-1}^2})\\
		&\propto \exp[-\frac{tr(\bar{\beta}_{t-1}^2\alpha_t^2+\beta_t^2)(\log \tilde{\mu}^{-\frac{1}{2}}\,X_0\,\tilde{\mu}^{-\frac{1}{2}})^2}{2\bar{\beta}_{t-1}^2\beta_t^2}]\\
		& \sim G(\tilde{\mu}(X_0,\epsilon),\tilde{\sigma}_t^2),\\
		& \mathrm{with} \quad \tilde{\mu}(x_0,\epsilon) = \frac{1}{\alpha_t} \odot X_t \ominus\frac{\beta_t^2}{\alpha_t \bar{\beta}_t} \odot \epsilon, \tilde{\sigma}_t = \frac{\bar{\beta}_{t-1}\beta_t}{\bar{\beta}_t}.
	\end{split} \label{ddpm_spd_q(xt_1x0)}
\end{equation}

Note that the final computation expression of $q(X_{t-1}|X_t, X_0)$ is similar to DDPM (\emph{i.e.}, Eq.~\ref{ddpm_eud_q(xt)}), while their difference is that our addition and scalar multiplication are defined through exponential and logarithmic mappings.
In Euclidean space, the exponential and logarithmic mappings can be regarded as addition and subtraction operations. 

Further, we also formulate $p(X_{t-1}|X_t)$ as:
\begin{equation}
\small
\setlength{\abovedisplayskip}{3pt}   
\setlength{\belowdisplayskip}{3pt} 
  {p(X_{t-1}|X_t) \sim G(\mu_{\theta}(X_t,t),\Sigma_t^2)},
\label{backward}
\end{equation}
where $\Sigma_t^2 = \beta_t^2$.  $\mu_{\theta}(X_t,t)$ can be computed according to $q(X_{t-1}|X_t,X_0)$:
\begin{equation}
\small
\setlength{\abovedisplayskip}{3pt}   
\setlength{\belowdisplayskip}{3pt} 
	\begin{split}
		\mu_{\theta}(X_t, t) = \frac{1}{\alpha_t} \odot X_t \ominus\frac{\beta_t^2}{\alpha_t \bar{\beta}_t} \odot \epsilon_{\theta}.
	\end{split}  \label{ddpm_spd_backward_mean}
\end{equation}

During training, we aim to minimize the KL divergence between $p(X_{t-1}|X_t)$ and $q(X_{t-1}|X_t, X_0)$. We also found that $\operatorname{KL}(q(X_{t-1}|X_t,X_0) \parallel p(X_{t-1}|X_t))$ is bounded by $d(\mu_{X_0,\epsilon},\mu_{\theta})^2$, \emph{i.e.},
\begin{equation}
\small
\setlength{\abovedisplayskip}{3pt}   
\setlength{\belowdisplayskip}{3pt} 
	\begin{split}
		&KL(q(X_{t-1}|X_t,X_0) \parallel p(X_{t-1}|X_t)) \\
		&\propto \frac{2\beta_t^2d(\mu_{\theta},X)^2-2\tilde{\sigma}_t^2d(\mu_{X_0,\epsilon},X)}{2\beta_t^2\tilde{\sigma}_t^2}\\
		& \propto \frac{2\beta_t^2[d(\mu_{\theta},X)^2-d(\mu_{x_0,\epsilon},X)^2]}{2\beta_t^2\tilde{\sigma}_t^2}\\
		& \le \kappa \,d(\mu_{X_0,\epsilon},\mu_{\theta})^2,
  \label{kl}
\end{split}  
\end{equation}
where $\kappa$ is a constant, and $d(\mu_{X_0,\epsilon},\mu_{\theta})^2$ is further simplified as:
\begin{equation}
\small
\setlength{\abovedisplayskip}{3pt}   
\setlength{\belowdisplayskip}{3pt} 
	\begin{split}
		&d(\mu_{X_0,\epsilon},\mu_{\theta})^2 \\
		&= d(\frac{1}{\alpha_t} \odot X_t \ominus \frac{\beta_t^2}{\alpha_t \bar{\beta}_t} \odot \epsilon,\,\frac{1}{\alpha_t} \odot X_t \ominus\frac{\beta_t^2}{\alpha_t \bar{\beta}_t} \odot \epsilon_{\theta})^2\\
		&= d(X_t^{-\frac{1}{2\alpha_t}} \epsilon^\frac{\beta_t^2}{\alpha_t \bar{\beta}_t} X_t^{-\frac{1}{2\alpha_t}},X_t^{-\frac{1}{2\alpha_t}} \epsilon_{\theta}^\frac{\beta_t^2}{\alpha_t \bar{\beta}_t} X_t^{-\frac{1}{2\alpha_t}})^2\\
		&= d(\epsilon^\frac{\beta_t^2}{\alpha_t \bar{\beta}_t},\epsilon_{\theta}^\frac{\beta_t^2}{\alpha_t \bar{\beta}_t})^2  \\  
		&= (\frac{\beta_t^2}{\alpha_t \bar{\beta}_t})^2d(\epsilon,\epsilon_{\theta})^2. 	
	\end{split}\label{ddpm_spd_kl2}
\end{equation}

For computational simplicity, we use $d(\epsilon,\epsilon_{\theta})^2$ as the training objective function. If we degenerate it to the Euclidean space, it will be $L_2$ distance to train DDPM. The detailed training process of SPD-DDPM is presented in Alg.~\ref{ddpm:uncon_train}.

For inference, we start from the standard Gaussian distribution in SPD space and gradually denoise the sample to generate $\hat{X}_0$, whose process is presented in Alg.~\ref{ddpm:uncon_sample}. $\gamma$ is used to balance the diversity and robustness. A smaller $\gamma$ leads to greater diversity in the generated matrix, but it may result in low quality. On the other hand, a larger $\gamma$ produces higher quality outputs, but with reduced diversity. 

Alg.~\ref{ddpm:uncon_train}, and ~\ref{ddpm:uncon_sample} present the unconditional training of SPD-DDPM, while our SPD-DDPM also works well under the conditional generation. In the following, we will first describe the Symmetric Positive Definite U-Net to fit $\epsilon$ and then propose the conditional process of SPD-DDPM.

\begin{algorithm}[t]
\small
	\caption{Unconditional Training of SPD-DDPM.}
	\label{ddpm:uncon_train}
	\begin{algorithmic}[1]
		\REPEAT{
			\STATE	$t \sim \text{Uniform}({1,\cdots, T})$
			\STATE   $X_0 \sim q(X_0)$
			\STATE  $\epsilon \sim G(I,1)$
			\STATE  $X_t = \bar{\alpha}_t \odot X_0 \oplus \bar{\beta}_t \odot \epsilon$
			\STATE  Take gradient descent step on\\
					$ \nabla_{\theta} \parallel \epsilon - \epsilon_{\theta}(X_t,t) \parallel $
		}
		\UNTIL{convergence}	
	\end{algorithmic}
\end{algorithm}

\begin{algorithm}[t]
\small
	\caption{Unconditional Sampling of SPD-DDPM.}
	\label{ddpm:uncon_sample}
	\begin{algorithmic}[1]
		\STATE $X_T \sim G(I,1)$
		\FOR{$t = T, \dots 1$}
			\STATE $z \sim G(I,1)$
			\STATE $X_{t-1} = \frac{1}{\alpha_t}\odot(X_t \ominus \frac{\beta_t^2}{\bar{\beta}_t}\epsilon_{\theta})\oplus \tilde{\sigma_t}/\gamma \odot z$, where $\ominus$ is defined by using $-$ to replace $+$ in Eq.~\ref{plus_mul_in_spd}.
		\ENDFOR
		\RETURN $X_0$
	\end{algorithmic}
\end{algorithm}

The detailed proofs presented in this section can be found in the supplement.

\subsection{Symmetric Positive Definite U-Net}
In SPD-DDPM, a neural network $\epsilon_{\theta}$ takes an SPD matrix and time $t$ as the inputs to output an SPD matrix.
Traditional U-Net in DDPM can not guarantee the symmetry and positive definiteness of the output.
Huang~\textit{et al.}~\cite{huang2017riemannian} proposed a simple SPD Net by defining the convolution and activation layers. However, the objective of this network framework is to transform the SPD matrix into a vector in Euclidean space, and this network has a limited capacity to fit $\epsilon$. Moreover, the above SPD Net can not incorporate the conditional factors $t$. It fails to adapt to the different distributions of $X_t$ at different moments.

In this paper, we propose a novel high-capacity SPD U-Net framework to estimate the SPD matrix.
\begin{equation}
\small
\setlength{\abovedisplayskip}{3pt}   
\setlength{\belowdisplayskip}{3pt} 
		X_k = f_b^{(k)}(X_{k-1};W_k) = W_k X_{k-1} W_k^{'},
\label{convolution}
\end{equation}
\begin{equation}
\small
\setlength{\abovedisplayskip}{3pt}   
\setlength{\belowdisplayskip}{3pt} 
		X_k = f_r^{(k)}(X_{k-1}) = U_{k-1}\max(\epsilon I,\Sigma_{k-1})U^{'}_{k-1}.
\label{activation}
\end{equation}

\begin{figure}
	\centering
	\includegraphics[width=0.9\linewidth]{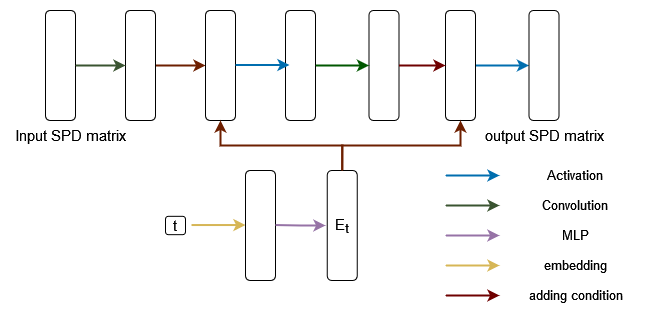}
	\caption{Double Convolution in SPD U-Net. Input matrix passes through convolution layers, followed by adding the condition of time factor $t$, and then activation layers. Repeat the above process making the double convolution block.}
	\label{fig:doublecon}
\end{figure}

Furthermore, we add the time factor $t$ by first putting it through an embedding layer following a shallow neural network to generate a matrix $E_t$. As shown in Fig.~\ref{fig:doublecon}, we incorporate the time factor $t$ into the neural network using the following equation:
\begin{equation}
\small
\setlength{\abovedisplayskip}{3pt}   
\setlength{\belowdisplayskip}{3pt} 
X_k  =  E_t X_{k-1} E_t^{'}.
\label{convolution_time}
\end{equation}

To better estimate the distribution of source data, we increase the network depth by adding double convolution as shown in Fig.~\ref{fig:doublecon}. The SPD matrix passes through a convolution layer using~\ref{convolution} and then adds condition $t$ by~\ref{convolution_time}. After that, we do activation using~\ref{activation}. We repeat the above computation operation once more as a basic unit, termed double convolution, which increases the network depth and enhances its fitting capability.

\begin{figure}[h]
	\centering
	\includegraphics[width=0.99\linewidth]{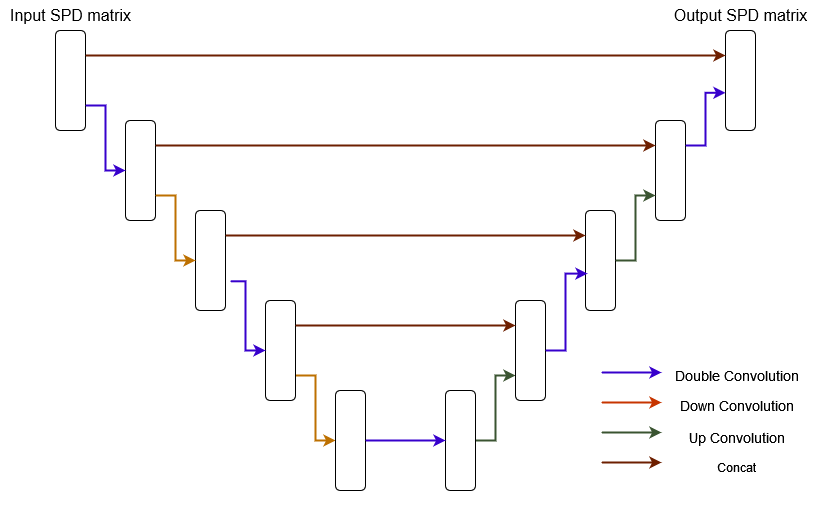}
	\caption{Architecture of SPD U-Net. SPD U-Net contains five double convolutions, two down convolution and two up convolution layers.}
	\label{fig:spdnet}
\end{figure}

The overall framework of our network is shown in Fig.~\ref{fig:spdnet}. The down convolution is defined as $W_k \in R^{d_k \times d_{k-1}}(d_k < d_{k-1})$. The up convolution is defined as filling the diagonal of the input matrix with $1$ to match the output size, and then performing convolution \eqref{convolution}. The concatenation layer is defined as computing the mean of two layers.

Different from Wang~\emph{et al.}~\cite{wang2023u}, our neural network is deeper and allows for conditional additions. Additionally, the objective of our network is not to reconstruct the input SPD matrix but to fit other SPD matrices instead.
We will conduct ablation experiments to evaluate the effectiveness of each component.

\subsection{Conditional Symmetric Positive Definite Denoising Diffusion Probabilistic Model}
We employ the classifier-free framework in the conditional generation, which additionally adds the conditional vector $y$ in the SPD net. Similar to the incorporation process of the time factor $t$, the conditional vector $t$ is transformed into a matrix via an embedding layer with a shallow network. Thus,
the training loss function in the conditional SPD-DDPM can be formulated as: 
\begin{equation}
\small
\setlength{\abovedisplayskip}{3pt}   
\setlength{\belowdisplayskip}{3pt} 
		\parallel\epsilon - \epsilon_{\theta}(X_t,t,y)\parallel.
\end{equation}
As presented in Alg.~\ref{ddpm:con_train}, the training process is similar to the unconditional SPD-DDPM. 

In conditional generation, we aim to incorporate conditional generation as a novel approach for regression modeling. In regression models, it is common to assume that $X$ given $y$ follows a certain distribution and use $E(X|y)$ as the prediction of $X$ given $y$. In this work, we assume $X|y \sim G(\mu_y,\sigma^2)$ with $\mu_y$ being $E(X|y)$. Said~\emph{et al.}~\cite{said2017riemannian} have deduced that the maximum likelihood estimation of $E(X|y)$ is the empirical Riemannian centre of samples from $P(X|y)$, where the Riemannian centre $\hat{X}_N$ is defined as:
\begin{equation}
\small
\setlength{\abovedisplayskip}{3pt}   
\setlength{\belowdisplayskip}{3pt} 
		\hat{X}_N = \arg\min\limits_X\frac{1}{N} \mathop{\sum}\limits_{n=1}^N d(X,X^n)^2 \label{MLE}.
\end{equation}

$\hat{X}_N$ under the affine-invariant metric in Eq.~\ref{MLE} can be obtained using a gradient descent algorithm~\cite{dryden2009non,pennec1999probabilities,pennec2006riemannian}.
Therefore, we train the model to fit $p(X|y)$. During inference, given a specific $y$, we first generate $N$ samples from $p(X|y)$. We then compute $\hat{X}_N$ by minimizing Eq.~\eqref{MLE} as the prediction of $E(X|y)$. The detailed inference process is presented in Alg.~\ref{ddpm:con_inference}. 

\begin{algorithm}[t]
\small
	\caption{Conditional Training of SDP-DDPM.}
	\label{ddpm:con_train}
	\begin{algorithmic}[1]
		\REPEAT{
			\STATE	$t \sim \text{Uniform}({1,\cdots, T})$
			\STATE  $X_0 \sim q(X_0)$
			\STATE  $\epsilon \sim G(I,1)$
			\STATE  $X_t = \bar{\alpha}_t \odot X_0 \oplus \bar{\beta}_t \odot \epsilon$
			\STATE  Take gradient descent step on\\
			$ \nabla_{\theta} \parallel \epsilon - \epsilon_{\theta}(X_t,t,y) \parallel $
		}
		\UNTIL{convergence}	
	\end{algorithmic}
\end{algorithm}

\begin{algorithm}[t]
\small
	\caption{Conditional Inference of SDP-DDPM.}
	\label{ddpm:con_inference}
	\begin{algorithmic}[1]
		\FOR{$i = 1,\dots, N$}
			\STATE $X_T^{i} \sim G(I,1)$
			\FOR{$t = T, \dots 1$}
			\STATE $z \sim G(I,1)$
			\STATE $X_{t-1}^{i} = \frac{1}{\alpha_t}\odot(X_t^{i} \ominus \frac{\beta_t^2}{\bar{\beta}_t}\epsilon_{\theta})\oplus \tilde{\sigma_t}/\gamma \odot z$
			\ENDFOR
		\ENDFOR
		\STATE $\hat{X} =  \arg\min\limits_X \frac{1}{N} \mathop{\sum}\limits_{n=1}^N d^2(X,X^n) $
		\RETURN $\hat{X}$
	\end{algorithmic}
\end{algorithm}

\section{Experiments}

\subsection{Experimental Setups}
\textbf{Datasets.}
We use the simulation or toy data to test the performance of unconditional SPD-DDPM. We first randomly select a matrix $A$ as the center point of the distribution, and then sample total $15,000$ SPD matrices from $G(A,\sigma^2)$ as the training data. 

In the conditional generation, taxi data from the New York City Taxi and Limousine Commission is selected, which  is available in \footnote{https://www.nyc.gov/site/tlc/about/tlc-trip-record-data.page.} Data preprocessing is following~\cite{tucker2023variable}. New York City is divided into $10$ boroughs (Detail refers to the supplementary materials). Each element of the SPD matrix $X_{ij}$ represents the passenger flow between the $i$-th borough and the $j$-th borough.
We obtain $8,700$ weighted adjacency matrices by collecting data from 2019.1 to 2019.12 and also collect the $13$ predictors following~\cite{tucker2023variable}. We select $7,600$ samples as the training set and $1,100$ samples as the testing set.

\textbf{Evaluation metric.}
In the unconditional generation, the Affine-invariant metric is widely used to test performance:
\begin{equation*}
\small
\setlength{\abovedisplayskip}{3pt}   
\setlength{\belowdisplayskip}{3pt} 
d(A,B)^2 = tr[\operatorname{log}(A^{-0.5}B A^{-0.5})]^2.
\end{equation*}
In the condition generation, there is no previous method that optimizes under the Affine-invariant metric, so we select both Frobenius and Affine-invariant metric as the evaluation protocol, where Frobenius distance is defined as:
\begin{equation*}
\small
\setlength{\abovedisplayskip}{3pt}   
\setlength{\belowdisplayskip}{3pt} 
d(A,B)^2 = \sqrt{\sum_{i=1}^m\sum_{j=1}^m(A_{ij}-B_{ij})^2}.
\end{equation*}

\textbf{Implementation details.}
We implement our method using PyTorch 2.0.1~\cite{paszke2017automatic} with one NVIDIA A6000 GPU. The models are trained using Adam optimizer~\cite{kingma2014adam} with default parameters to DDPM. The learning rate is initialized by 0.0015 decaying by each iteration with the cosine function. The batch size is set to 150 and 100, and the total epochs are set to 50 and 20 in the unconditional and conditional generation, respectively. We set $\alpha_t$ and $T$ to $\sqrt{1-\frac{0.08t}{T}}$ and 200, respectively. $\gamma$ in Alg.~\ref{ddpm:uncon_sample} and ~\ref{ddpm:con_inference} is both default set to 10 unless the specific discuss. Code is available at \url{https://github.com/li-yun-chen/SPD-DDPM}.

\subsection{Unconditional Generation}
We generate $300$ SPD matrices and use a heatmap to visualize the training samples and generated samples. Each heatplot shows the value of one SPD matrix. To evaluate the effectiveness of our unconditional SPD-DDPM, we compare our model with DDPM.

\begin{figure}
	\centering
	\includegraphics[scale = 0.23]{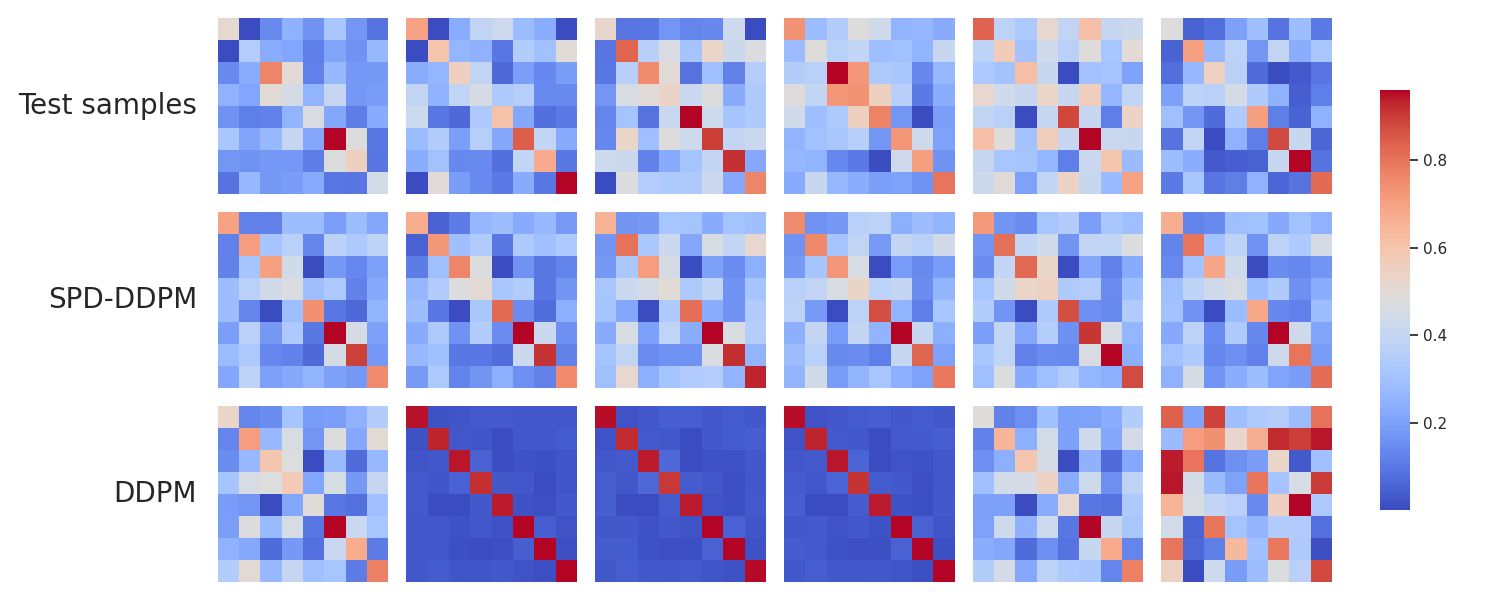}
	\caption{Visualization of SPD samples. We randomly select six samples from test samples, generated by SPD-DDPM and DDPM. Each column presents one sample.}
	\label{fig:uncondition}
\end{figure}

We first calculate the mean distance between the generated samples and the real distribution as a measure of evaluating the quality of generation.
The results are summarized in Tab.~\ref{tab:uncondition}. compared to DDPM, our unconditional SPD-DDPM achieves a lower mean distance. Furthermore, we visualize the generated samples in Fig.~\ref{fig:uncondition}. Obviously, our models are better to fit the data, compared to DDPM. 
In this experiment, our data in the SPD space is a low-dimensional manifold, directly using DDPM to generate samples cannot be effective for the generation. In contrast, our unconditional SPD-DDPM fully considers the SPD information for better generation.

\begin{table}[t]
\small
\setlength{\abovedisplayskip}{3pt}   
\setlength{\belowdisplayskip}{3pt} 
	\centering
	\caption{Mean distance between the generated samples and the real distribution in the unconditional generation. $a\pm b$ means the mean and std in all tables by running the model 5 times, respectively.}
	\begin{tabular}{c|c}
		\toprule
		Method & \multicolumn{1}{l}{Mean Distance$\downarrow$} \\
		\midrule
  DDPM  & 813.75$\pm$ 187.92 \\
		SPD-DDPM & 23.17$\pm$ 3.00 \\
		\bottomrule
	\end{tabular}%
	\label{tab:uncondition}%
\end{table}%

\subsection{Conditional Generation}



Our goal is to use predictor variables $y$ to generate responding SPD matrices, achieving generative predictive modeling. Practically, we generate $n=20$ samples for each predictor $y$. We compare our method with the discriminative method Frechet Regression~\cite{petersen2019frechet} and DDPM. Our method is optimized under the Affine-invariant metric, while Frechet Regression is optimized under the Forbinus metric. The results are summarized in Tab.~\ref{tab:con_dis}. Our method outperforms Frechet Regression in both Frobenius and Affine-invariant metrics. In the Frobenius metric, our method slightly outperforms Frechet Regression because it is optimized under the Frobenius norm. In the Affine-invariant metric, our method significantly outperforms Frechet Regression. Our conditional SPD-DDPM also achieves significantly lower Frobenius metric, compared to DDPM.

To further evaluate the effectiveness of our method, we visualize the SPD matrices in the Manhattan map Fig. \ref{fig:taxi}. The left visualization represents the Frechet Regression, followed by the visualizations for our method and real data. We found that the predictions of Frechet Regression are overestimated and fail to capture the differences between each edge effectively. While our method shows a closer approximation to the real matrix. 

Actually, discriminative methods require inputting all the data into the model simultaneously, rather than training in batches. When dealing with large-scale data, it will exceed memory limitations. Furthermore, their approach lacks the capability to retain intermediate model parameters, necessitating retraining the model when new samples need estimation. Our approach is based on the neural network, allowing us to circumvent the issues mentioned above.

\begin{figure}[t]
	\centering
	\includegraphics[scale=0.6]{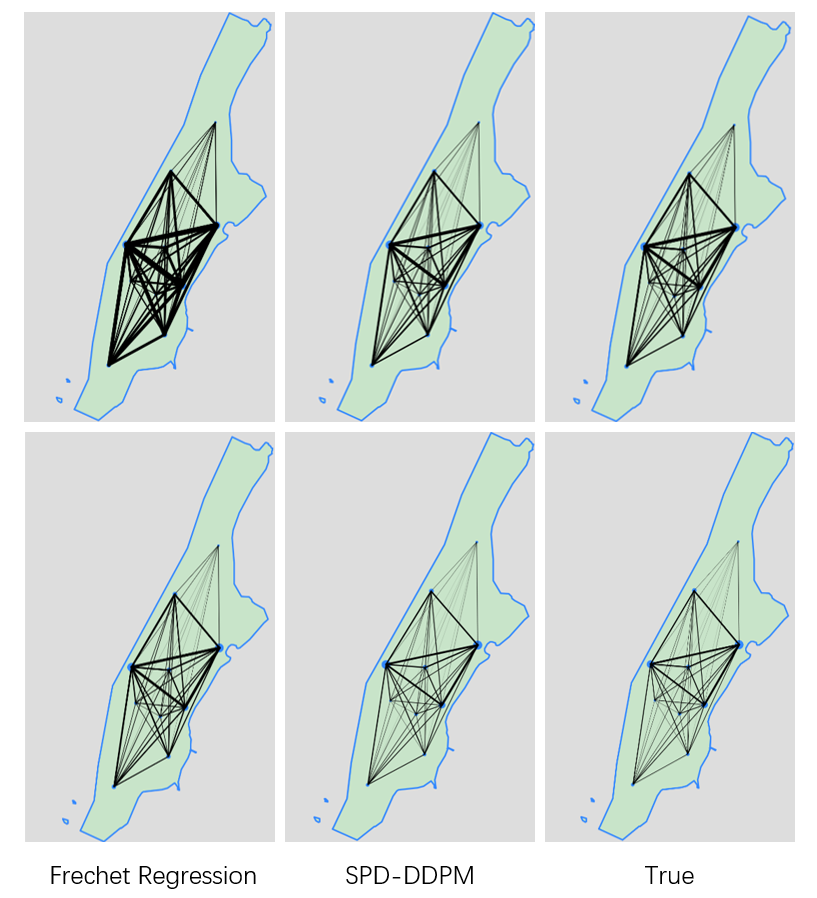}
	\caption{Visualization of taxi data. On the left is the data predicted by Frechet Regression, in the middle is the data predicted by our method, and on the right is the real data. Each node represents a region, and the thickness of the line segments represents the number of taxis.}
	\label{fig:taxi}
\end{figure}

\begin{table}[t]
\small
\setlength{\abovedisplayskip}{3pt}   
\setlength{\belowdisplayskip}{3pt} 
	\centering
	\caption{Comparsion among Frechet Regression, DDPM and SPD-DDPM for conditional generation.}
	\begin{tabular}{c|c|c}
		\toprule
		Method&{Frobenius $\downarrow$} & {Affine-invariant $\downarrow$} \\
		\midrule
		Frechet Regression& 6.6448     & 8.0099 \\
        DDPM & 62.97     & 1203.10 \\
        SPD-DDPM & 6.4684     & 0.6123 \\
		\bottomrule
	\end{tabular}%
	\label{tab:con_dis}%
\end{table}%



\subsection{Ablation Study}
In this ablation study, we mainly evaluate the effects of hyperparameter $\gamma$ and SPD U-Net, which are very important to SPD-DDPM.

\textbf{Effect of $\gamma$.} 
We evaluate the effect of $\gamma$ in unconditional generation. As shown in Tab. \ref{abla_r}, larger $\gamma$ generates a lower mean distance to better estimate the inputs. In addition, the std is also reduced by increasing $\gamma$. 

\begin{table}[t]
\small
\setlength{\abovedisplayskip}{3pt}   
\setlength{\belowdisplayskip}{3pt} 
	\centering
	\caption{Effect of $\gamma$ for unconditional generation. }
	\begin{tabular}{c|c}
		\toprule
		Method & \multicolumn{1}{l}{Mean Distance$\downarrow$} \\
		\midrule
		SPD-DDPM($\gamma$ = 2) & 65.81$\pm$ 17.86 \\
		SPD-DDPM($\gamma$ = 4) & 35.99$\pm$ 7.58 \\
		SPD-DDPM($\gamma$ = 10) & 23.17$\pm$ 3.00 \\
		\bottomrule
	\end{tabular}%
	\label{abla_r}%
\end{table}%

\textbf{Effect of SPD U-Net.} 
In this paper, we propose a high-capacity SPD U-Net. Compared to the traditional SPD net, the network proposed in this paper is deeper and incorporates a double convolution, down and up convolution, and concatenation structure. We further demonstrate the effectiveness of these modifications through ablation experiments in unconditional generation by comparing the mean distance between samples and the distribution mean under $\gamma = 10$.

\begin{table}[t]
\small
\setlength{\abovedisplayskip}{3pt}   
\setlength{\belowdisplayskip}{3pt} 
	\centering
	\caption{Ablation Study of SPD U-Net.}
	\begin{tabular}{c|c}
		\toprule
		Method & \multicolumn{1}{l}{Mean distance$\downarrow$} \\
		\midrule
		Without double convolution & 26.17$\pm$ 3.75 \\
		Without concatenation & -- \\
		Without down\&up convolution & 42.75$\pm$ 3.25 \\
		SPD-DDPM with SPD U-Net & 23.17$\pm$ 3.00 \\
		\bottomrule
	\end{tabular}%
	\label{tab:addlabel}%
\end{table}%

If we replace the double convolution with a single layer, reducing the number of convolutional layers by half, the performance of model generation decreases 3 mean distance. It indicates the effectiveness of double convolution for network's fitting capability. 
The generation will fails when removing the concatenation structure. Thus, concatenation operations play a vital role in our SPD U-Net. 
Removing the down and up convolutions, each layer uses the matrices with the same dimensions, and the distance will increase to $42.75$, which significantly affects the model's generalization ability.

\section{Related Work}

\subsection{Denoising diffusion probabilistic model}
DDPM is a generative model proposed by Ho~\textit{et al.}~\cite{ho2020denoising}.
Nichol~\textit{et al.}~\cite{nichol2021improved} and Song~\textit{et al.}~\cite{song2020denoising} further made improvements to it. Dhariwal~\textit{et al.}~\cite{dhariwal2021diffusion} proposed the classifier-guidance version, while Ho~\textit{et al.}~\cite{ho2022classifier} proposed the classifier-free version. The diffusion model is first applied in computer vision and found numerous applications in conditional image generation~\cite{nichol2021glide,rombach2022high}, image restoration~\cite{wang2022zero,lugmayr2022repaint}, 3D image generation~\cite{wang2023prolificdreamer}, and other fields.
Song~\textit{et al.}~\cite{song2020score} introduced the score-based generation method to accelerate the generation speed of the diffusion model from the perspective of stochastic differential equations~\cite{lu2022dpm}. Different with these methods, our SPD-DDPM is constructed on the SPD space, rather than Euclidean space. By introducing Gaussian distribution, the addition and multiplication operations in the SPD space, we effectively generate the forward and backward processes in the SPD space for better estimating the SPD matrix unconditionally and conditionally.

\subsection{Symmetric positive definite metric}
SPD space is a nonlinear metric space.
Depending on the metric, SPD forms a Riemannian manifold.
Different matrices have been proposed, such as affine-invariant metric~\cite{moakher2005differential,pennec2006intrinsic,fletcher2007riemannian}, log-Euclidean metric~\cite{arsigny2007geometric}, log-Cholesky metric~\cite{lin2019riemannian} and scaling rotation distance~\cite{jung2015scaling}.
Some work studied regression with SPD matrix valued responses.
Using Frechet mean~\cite{frechet1948elements}, Petersen~\textit{et al.}~\cite{petersen2019frechet} proposed Frechet Regression under different matric.
Based on this, Tucker~\textit{et al.}~\cite{tucker2023variable} proposed a method for variable selection.
Lin~\textit{et al.}~\cite{lin2023additive} proposed an adaptive model in SPD space.
Qiu~\textit{et al.}~\cite{qiu2022random} propose random forest with SPD matrix response. Different from these methods, our SPD-DDPM is the first generation model in SPD space, which can fit probability distribution in SPD space.

\section{Conclusion}

This paper proposes a novel denoising diffusion probabilistic model in the SPD space, termed SDP-DDPM.
The proposed SDP-DDPM is applied for both unconditional and conditional generation.
In the unconditional version, SDP-DDPM fits the probability distribution and generates samples in SPD space.
In the conditional version, the model generates the distribution of SPD given a specific condition and calculates the expectation $E(X|y)$ as the prediction.
A high-capacity SPD U-Net structure is further introduced to improve the data fitting performance.
Experimental results demonstrate the strong capability of the proposed method in the fitting probability distributions.

\section{Acknowledgements}

This paper is sponsored by the National Natural Science Foundation of China (NO. 62102151, NO. 12371289), the Shanghai Sailing Program (21YF1411200), Basic Research Project of Shanghai Science and Technology Commission(NO. 22JC1400800), CCF-Tencent Rhino-Bird Open Research Fund, the Open Research Fund of Key Laboratory of Advanced Theory and Application in Statistics and Data Science, Ministry of Education (KLATASDS2305), and the Fundamental Research Funds for the Central Universities.

\bibliography{aaai24}
\section{Operations in the SPD Space}
In this section, we will derive some operational properties in the SPD space, which will help us deduce properties of SPD-DDPM.

\subsection{Distributive Law of Multiplication}

Suppose $a$ is a scale, $X, Y$ are SPD matrices. Then 
\begin{equation}
\small
\setlength{\abovedisplayskip}{3pt} 
\setlength{\abovedisplayskip}{3pt} 
	a \odot (X \oplus Y) = a \odot X \oplus a \odot Y.\label{distribute law}
\end{equation}
Proof:
\begin{equation*}
\small
\setlength{\abovedisplayskip}{3pt} 
\setlength{\abovedisplayskip}{3pt} 
	\begin{split}
		a \odot (X \oplus Y) &= a \odot \exp(\log X + \log Y)\\
		&= \exp[a \cdot \log(\exp(\log X + \log Y))]\\
		&= \exp[a \cdot(\log X + \log Y)]\\
		&= \exp[a \log X + a \log Y]\\
		&= \exp\{\log[\exp(a \log X)] + \log[ \exp(a \log Y)]\}\\
		&= \exp(a \log X)\oplus \exp(a \log Y)\\
		&= a \odot X \oplus a \odot Y. 
	\end{split} 
\end{equation*}

\subsection{Combining Similar Terms}

Suppose $a,b$ is a scale, $X$ is an SPD matrix. Then 

\begin{equation}
\small
\setlength{\abovedisplayskip}{3pt} 
\setlength{\abovedisplayskip}{3pt} 
	a \odot X \oplus b \odot X = (a+b)\odot X.\label{conbine}
\end{equation}
Proof: 
\begin{equation*}
\small
\setlength{\abovedisplayskip}{3pt} 
\setlength{\abovedisplayskip}{3pt} 
	\begin{split}
		&a \odot X \oplus b \odot X \\
		&= \exp(a \log X) \oplus \exp(b \log X)\\
		&= \exp\{\log[\exp(a\log X)]+\log[\exp(b\log X)]\}\\
		&= \exp[a\log X+b\log X]\\
		&= \exp[(a+b)\log X]\\
		&= (a+b) \odot X.  
	\end{split}
\end{equation*}

\subsection{Multiplication of Gaussian Distribution in the SPD Space}

Suppose $a$ is a scale, $\epsilon \sim G(I,\sigma^2)$, Then  
\begin{equation}
\small
\setlength{\abovedisplayskip}{3pt} 
\setlength{\abovedisplayskip}{3pt} 
	a \odot \epsilon \sim G(I,a^2\sigma^2). \label{multiplication}
\end{equation}
Proof:

\( d(\epsilon^a,I)^2 = tr[\log\epsilon^a]^2 = tr[a \log \epsilon]^2 = a^2tr[\log \epsilon]^2 \)
\begin{equation*}
\small
\setlength{\abovedisplayskip}{3pt} 
\setlength{\abovedisplayskip}{3pt} 
	\begin{split}
		P(d(\epsilon^a,I)^2<a^2) &=P(d(\epsilon,I)^2<(\frac{r}{a})^2)\\
		&= \frac{\int_0^{\frac{r^2}{a^2}}e^{-\frac{u}{\sigma_t^2}}\,du}{\int_0^{\inf}e^{-\frac{u}{\sigma^2}}\,dz}\\
		&= \frac{\int_0^{r^2}e^{-\frac{j}{\sigma^2a^2}}\,dj}{\int_0^{\inf}e^{-\frac{z}{\sigma^2a^2}}\,dz}\\
		& \sim G(I,a^2\sigma^2). 
	\end{split}
\end{equation*}

\subsection{Constant Plus of Gaussian Distribution in the SPD Space}
Suppose $Y$ is a constant SPD matrix, $\epsilon \sim G(I,\sigma^2)$. Then 
\begin{equation}
small
\setlength{\abovedisplayskip}{3pt} 
\setlength{\abovedisplayskip}{3pt} 
	Y \oplus \epsilon \sim G(Y, \sigma^2).\label{constant plus} 
\end{equation}
Proof:
\begin{equation*}
\small
\setlength{\abovedisplayskip}{3pt} 
\setlength{\abovedisplayskip}{3pt} 
	\begin{split}
		Y \oplus \epsilon &= \exp[\log Y+\log\epsilon]\\
		&= \exp[\frac{1}{2}\log Y + \log\epsilon +\frac{1}{2}\log Y] \\
		&= \exp[\log(Y^{\frac{1}{2}}\,\epsilon \,Y^{\frac{1}{2}})]\\
		&= Y^{\frac{1}{2}}\,\epsilon\, Y^{\frac{1}{2}}\\
		& \sim G(Y,\sigma^2).
	\end{split}
\end{equation*}

\subsection{Plus of Gaussian Distribution in the SPD Space}
Suppose $\epsilon_1 \sim G(I,\sigma_1^2)$, $\epsilon_2 \sim G(I,\sigma_2^2)$. Then
\begin{equation}
small
\setlength{\abovedisplayskip}{3pt} 
\setlength{\abovedisplayskip}{3pt} 
	\epsilon_1 \oplus \epsilon_2 \sim G(I,\sigma_1^2+\sigma_2^2). \label{plus of gussian distribution}
\end{equation}
Proof:

Define $Z = \epsilon_1 \oplus \epsilon_2$
\begin{equation*}
\small
\setlength{\abovedisplayskip}{3pt} 
\setlength{\abovedisplayskip}{3pt} 
	\begin{split}
		&P(Z = a) \\
		&= \int_{P_m} p(\epsilon_1 = b)p(\epsilon_2 = a\ominus b ) \,dv(b)\\
		&= \int_{P_m} \frac{1}{\zeta(\sigma_1)}\frac{1}{\zeta(\sigma_2)} \exp[-\frac{d(I,b)^2}{2\sigma_1^2}-\frac{d(I,a \ominus b)^2}{2 \sigma_2^2}] \,dv(b)\\
		&= \int_{P_m} \frac{1}{\zeta(\sigma_1)} \frac{1}{\zeta({\sigma_2})} \exp[\\
		&\qquad -\frac{(\sigma_1^2+\sigma_2^2)tr[\log B]^2}{2\sigma_1^2 \sigma_2^2}\\
		&\qquad +\frac{2\sigma_1^2tr[\log a \log b]-\sigma_1^2tr[\log a]^2}{2\sigma_1^2 \sigma_2^2}]\,dv(b)\\
		&= \int_{P_m} \frac{1}{\zeta(\sigma_1)} \frac{1}{\zeta({\sigma_2})} \exp[\\
		&\qquad -\frac{tr[\sqrt{\sigma_1^2+\sigma_2^2}\log b-\frac{\sigma_1^2}{\sqrt{\sigma_1^2+\sigma_2^2}}\log a]^2}{2\sigma_1^2\sigma_2^2}\\
		&\qquad+\frac{[\sigma_1^2-\frac{\sigma_1^4}{\sigma_1^2+\sigma_1^2}]tr[\log a]^2}{2\sigma_1^2\sigma_2^2}]\,dv(b)\\
		&= \exp[-\frac{1-\frac{\sigma_1^2}{\sigma_1^2+\sigma_2^2}}{2\sigma_2^2}tr[\log a]^2]\frac{1}{\zeta(\sigma_1)} \frac{1}{\zeta(\sigma_2)} \\
		&\qquad\int_{P_m}  \exp[-\frac{tr[\sqrt{\sigma_1^2+\sigma_2^2}\log b-\frac{\sigma_1}{\sqrt{\sigma_1^2+\sigma_2^2}}\log a]^2}{2\sigma_1^2\sigma_2^2}]dv(b)\\
		&\propto \, \exp[-\frac{d(I,a)^2}{2(\sigma_1^2+\sigma_2^2)}].
	\end{split}
\end{equation*}

\subsection{Distance of matrix power}
Suppose $a$ is a scale, $X$ and $Y$ are SPD matrices. Then 

\begin{equation}
\small
\setlength{\abovedisplayskip}{3pt} 
\setlength{\abovedisplayskip}{3pt} 
	d(X^a,Y^a)^2 = a^2\,d(X,Y)^2. \label{distance of matrix power}
\end{equation}
Proof:
\begin{equation*}
\small
\setlength{\abovedisplayskip}{3pt} 
\setlength{\abovedisplayskip}{3pt} 
	\begin{split}
		d(X^a,Y^a)^2 &= tr[\log X^{-\frac{a}{2}}B^aY^{-\frac{a}{2}}]^2\\
		&= tr[-\frac{a}{2}\log X \oplus a\log Y \oplus-\frac{a}{2}\log X]^2\\
		&= a^2tr[-\frac{1}{2}\log X \oplus \log Y \oplus-\frac{1}{2}\log X]^2\\
		&= a^2 tr[\log X^{-\frac{1}{2}}YX^{-\frac{1a}{2}}]^2\\
		&= a^2\,d(X,Y)^2.
	\end{split} 
\end{equation*}

\section{Reparameterization of $q(X_t|X_0)$}
If $q(X_t|X_{t-1}) \sim G(\alpha_t \odot X_{t-1}, \beta_t^2)$,  $X_t$ can be written as:
\begin{equation*}
small
\setlength{\abovedisplayskip}{3pt} 
\setlength{\abovedisplayskip}{3pt} 
	\begin{split}
		X_t = \alpha_t \odot X_{t-1} \oplus \beta_t \odot \epsilon_t, \;\epsilon_t \sim G(I,1).
	\end{split}
\end{equation*}
This is further promoted as:
\begin{equation}
\small
\setlength{\abovedisplayskip}{3pt} 
\setlength{\abovedisplayskip}{3pt} 
	\begin{split}
		X_t &= \alpha_t \odot X_{t-1} \oplus \beta_t \odot \epsilon_t\\
		&= \alpha_t \odot (\alpha_{t-1} \odot X_{t-2} \oplus \beta_{t-1}\epsilon_{t-1})\oplus \beta_t \odot \epsilon_t\\
		&= \alpha_t \alpha_{t-1}\odot X_{t-2} \oplus \alpha_t \beta_{t-1} \odot \epsilon_{t-1} \oplus \beta_t \odot \epsilon_t \quad \mathrm{follow} \;\eqref{distribute law} \\
		&= (\alpha_t\dots \alpha_1)\odot X_0 \oplus (\alpha_t \dots \alpha_2)\beta_1 \epsilon_1 \oplus \dots \oplus \beta_t\epsilon_t\\
		&= \bar{\alpha}_t \odot X_0 \oplus\sqrt{1-\bar{\alpha}_t^2} \odot \bar{\epsilon}_t, \, \bar{\epsilon}_t \sim G(I,1) \quad \mathrm{follow} \;\eqref{plus of gussian distribution}\\
		&\sim G(\bar{\alpha}_t \odot X_0 ,1-\bar{\alpha}_t^2). \quad \mathrm{follow} \;\eqref{constant plus} \eqref{multiplication}
	\end{split} \label{x_tx_0}
\end{equation}

\section{Computation of $q(X_{t-1}|X_t,X_0)$}

\begin{equation*}
\small
\setlength{\abovedisplayskip}{3pt} 
\setlength{\abovedisplayskip}{3pt} 
	\begin{split}
		&q(X_t|X_{t-1},X_0) \\
		&= q(X_t|X_{t-1})\frac{q(X_{t-1}|X_0)}{q(X_t|X_0)}\\
		&= G(\alpha_t \odot X_{t-1},\beta_t) \frac{G(\bar{\alpha}_{t-1} \odot X_0, \bar{\beta}_{t-1})}{G(\bar{\alpha}_{t} \odot x_0, \bar{\beta}_{t})}\\
		& = \frac{\zeta(\bar{\beta}_t)}{\zeta(\beta_t)\zeta(\bar{\beta}_{t-1})} \exp(-\frac{d(X_t,X_{t-1}^{\alpha_{t}})^2}{2 \beta_t^2} \\
		&\qquad -\frac{d(X_{t-1},x_0^{\bar{\alpha}_{t-1}})}{2\bar{\beta}_{t-1}^2}+\frac{d(X_{t},x_0^{\bar{\alpha}_{t}})}{2\bar{\beta}_{t^2}})\\
		& \propto \exp(-\frac{d(X_t,X_{t-1}^{\alpha_{t}})^2}{2 \beta_t^2}-\frac{d(X_{t-1},X_0^{\bar{\alpha}_{t-1}})}{2\bar{\beta}_{t-1}^2})\\
		& = \exp(-\frac{\bar{\beta}_{t-1}^2 d(X_t,X_{t-1}^{\alpha_{t}})^2+\beta_{t}^2 d(X_{t-1},X_0^{\bar{\alpha}_{t-1}})}{2\beta_t^2 \bar{\beta}_{t-1}^2})\\
		& = \exp(-\frac{\bar{\beta}_{t-1} tr[\log(X_t^{-\frac{1}{2}}X_{t-1}^{\alpha_t}X_t^{-\frac{1}{2}})]^2}{2\beta_t^2 \bar{\beta}_{t-1}^2}-\\
		&\qquad\frac{\beta_{t}^2tr[\log( X_0^{-\frac{\bar{\alpha}_{t-1}}{2}} X_{t-1} X_0^{-\frac{\bar{\alpha}_{t-1}}{2}})]^2}{2\beta_t^2 \bar{\beta}_{t-1}^2})\\
		& = \exp(\frac{tr[\bar{\beta}_{t-1}^2 \log(X_t^{-\frac{1}{2}}X_{t-1}^{\alpha_t}X_t^{-\frac{1}{2}})}{2\beta_t^2 \bar{\beta}_{t-1}^2}-\\
		&\qquad	\frac{\beta_{t}^2\log( X_0^{-\frac{\bar{\alpha}_{t-1}}{2}} X_{t-1} X_0^{-\frac{\bar{\alpha}_{t-1}}{2}})]^2}{2\beta_t^2 \bar{\beta}_{t-1}^2}).
	\end{split}
\end{equation*}
Translate $A = X_t^{-\frac{1}{2}}$ ,$B = X_{t-1}$, $C = X_0^{-\frac{\bar{\alpha}_{t-1}}{2}}$, $a = \bar{\beta}_{t-1}$, $b = \beta_t$, $r = \alpha_t$.

Then the numerator is written as:
\begin{equation}
\small
\setlength{\abovedisplayskip}{3pt} 
\setlength{\abovedisplayskip}{3pt} 
	\begin{split}
		 &a^2(\log AB^rA)^2 + b^2(\log CBC)^2 \\
		 & = a^2(\log A+\log B^r+\log A)^2 \\
		 & \qquad +b^2(\log C+\log B+\log C)^2\\
		 &= 4a^2(\log A)^2 +a^2(\log B^r)^2 +2a^2\log A\log B^r\\
		 & \qquad+2a^2\log B^r\log A +4b^2(\log C)^2+b^2(\log B)^2\\
		 & \qquad +2b^2\log C\log B+2b^2\log B\log C\\
		 &= (a^2r^2+b^2)(\log B)^2 +2(a^2r \log A+b^2 \log C)\log B \\
		 &\quad+2\log B(a^2 r\log A+b^2 \log C)+4a^2(\log A)^2 +4b^2(\log C)^2\\
		 &= (a^2r^2+b^2)[(\log B)^2 + 2\frac{a^2r\log A+b^2\log C}{a^2r^2+b^2}\log B\\
		 &\quad+2\log B\frac{a^2r\log A+b^2\log C}{a^2r^2+b^2}]+4a^2(\log A)^2+4b^2(\log C)^2\\
		 & \propto(a^2r^2+b^2)(\log D+\log B+\log D)^2,\\
		 &\qquad \log D = \frac{a^2r\log A+b^2\log C}{a^2r^2+b^2}\\
		 &= (a^2r^2+b^2)(\log DBD)^2,\\
		 &\qquad D = A^{\frac{a^2r}{2(a^2r^2+b^2)}}C^{\frac{b^2}{a^2r^2+b^2}}A^{\frac{a^2r}{2(a^2r^2+b^2)}}.	 
	\end{split}
\end{equation}
Thus $q(X_{t-1}|X_t,X_0)$ is written as:
\begin{equation}
\small
\setlength{\abovedisplayskip}{3pt} 
\setlength{\abovedisplayskip}{3pt} 
	\begin{split}
		&q(X_{t-1}|X_t,X_0) \\
        &\propto \exp[-\frac{tr(a^2r^2+b^2)(\log DBD)^2}{2\bar{\beta}_{t-1}^2\beta_t^2}]\\
		&= \exp[-\frac{tr[\log DBD]^2}{2\frac{\bar{\beta}_{t-1}^2\beta_t^2}{a^2r^2+b^2}}]\\
		&\sim G(\mu_t , \tilde{\sigma}_{t-1}^2),\\
        &\tilde{\sigma}_{t-1}^2 =\frac{\bar{\beta}_{t-1}^2\beta_t^2}{a^2r^2+b^2} = \frac{\bar{\beta}_{t-1}^2\beta_t^2}{\bar{\beta}_{t-1}^2\alpha_t^2+\beta_t^2} = \frac{\bar{\beta}_{t-1}^2\beta_t^2}{\bar{\beta}_t^2},
	\end{split}
\end{equation}
\begin{equation*}
\small
\setlength{\abovedisplayskip}{3pt} 
\setlength{\abovedisplayskip}{3pt} 
	\begin{split}
		\mu_t &= D^{-2} \\
		&=A^{-\frac{a^2r}{2(a^2r^2+b)}}C^{-\frac{b}{2^2(a^2r^2+b)}}A^{-\frac{a^2r}{2(a^2r^2+b)}}\\
		& \quad C^{-\frac{b}{2^2(a^2r^2+b)}}A^{-\frac{a^2r}{2(a^2r^2+b)}}\\
		&= X_t^{\frac{\bar{\beta}_{t-1}^2 \alpha_t}{4(\alpha_t^2\bar{\beta}^2_{t-1}+\beta_t^2)}} X_0^{\frac{\bar{\alpha}_{t-1}\beta_t^2}{2(\alpha_t^2\bar{\beta}^2_{t-1}+\beta_t^2)}}X_t^{\frac{\bar{\beta}_{t-1}^2 \alpha_t}{4(\alpha_t^2\bar{\beta}^2_{t-1}+\beta_t^2)}}\\ 
		& \quad X_0^{\frac{\bar{\alpha}_{t-1}\beta_t^2}{2(\alpha_t^2\bar{\beta}^2_{t-1}+\beta_t^2)}}X_t^{\frac{\bar{\beta}_{t-1}^2 \alpha_t}{4(\alpha_t^2\bar{\beta}^2_{t-1}+\beta_t^2)}}\\
		&= \frac{\bar{\beta}_{t-1}^2 \alpha_t}{(\alpha_t^2\bar{\beta}^2_{t-1}+\beta_t^2)} \odot X_t \oplus \frac{\bar{\alpha}_{t-1}\beta_t^2}{(\alpha_t^2\bar{\beta}^2_{t-1}+\beta_t^2)} \odot X_0\\
		&= \frac{\bar{\beta}_{t-1}^2 \alpha_t}{(\alpha_t^2\bar{\beta}^2_{t-1}+\beta_t^2)} \odot X_t \\
		&\quad \oplus \frac{\bar{\alpha}_{t-1}\beta_t^2}{(\alpha_t^2\bar{\beta}^2_{t-1}+\beta_t^2)} \odot [\frac{1}{\bar{\alpha}_t}(X_t \ominus \sqrt{\alpha_t^2\bar{\beta}^2_{t-1}+\beta_t^2}\epsilon_t]\\
		&= \frac{\beta_t^2+\alpha_t^2\bar{\beta}^2_{t-1}}{\alpha_t(\alpha_{t}^2\bar{\beta}_{t-1}^2+\beta_t^2)} \odot X_t \ominus \frac{\beta_t^2}{\alpha_t \sqrt{\alpha_{t}^2\bar{\beta}_{t-1}^2+\beta_t^2}} \odot \epsilon_t\\
		&= \frac{1}{\alpha_t} \odot X_t \ominus\frac{\beta_t^2}{\alpha_t \bar{\beta}_t} \odot \epsilon_t.
	\end{split}
\end{equation*}

\section{Upper Bound of KL divergence}
Suppose
\begin{equation*}
\small
\setlength{\abovedisplayskip}{3pt} 
\setlength{\abovedisplayskip}{3pt} 
	\begin{split}
		&q(X) \sim G(\mu_1,\sigma_1^2)\\
		&p(X) \sim G(\mu_2, \sigma_2^2).
	\end{split}
\end{equation*}
Then their KL divergence is bounded by:

\begin{equation}
\small
\setlength{\abovedisplayskip}{3pt} 
\setlength{\abovedisplayskip}{3pt} 
	\begin{split}
		&KL(q(X) \parallel p(X)) \\
		&\propto \frac{2\sigma_1^2d(\mu_2,X)^2-2\sigma_2^2d(\mu_1,X)}{2\sigma_1^2\sigma_2^2}\\
		& \propto \frac{2\sigma_1^2[d(\mu_2,X)^2-d(\mu_1,X)^2]}{2\sigma_1^2\sigma_2^2}\\
		& \le \kappa \,d(\mu_1,\mu_2)^2,\label{kl}
	\end{split}
\end{equation}
where $\kappa$ is a constant.

\section{Addition material for taxi data}
\subsection{Predictor}
The predictior in experiment condition SPD-DDPM are as follows.
\begin{itemize}
	\item Distance: Mean distance traveled, standardized
    \item Fare: Mean fare, standardized
	\item Passengers: Mean number of passengers, standardized
    \item Cash: Sum of cash indicators for type of payment, standardized
	\item Credit: Sum of credit indicators for type of payment, standardized
	\item Dispute: Sum of dispute indicators for type of payment, standardized
    \item Free: Sum of free indicators for type of payment, standardized
	\item Late Hour: Indicator for the hour being between 11pm and 5am, standardized
	\item average temperature
	\item average humidity
    \item average windspeed
    \item average pressure
    \item precipitation
\end{itemize}

\subsection{Division of NewYork}
In condition generation, Manhattan is divided into 10 zones, the division is described in Tab. \ref{tab:towns}

\begin{table}[htbp]
  \centering
  \caption{10 zones in Manhattan defined for the New York taxi data analysis}
    \begin{tabular}{|c|c|}
    \hline
    Zone & Town \\
    \hline
    1     & Inwood, Fort George, Washington Heights, \\
          &  Hamilton Heights, Harlem, East Harlem \\
    2     & Upper West Side, Morningside Heights, Central Park  \\
    3     & Yorkville, Lenox Hill, Upper East Side  \\
    4     & Lincoln Square, Clinton, Chelsea, Hell’s Kitchen  \\
    5     & Garment District, Theatre District  \\
    6     & Midtown  \\
    7     & Midtown South \\
    8     & Turtle Bay, Murray Hill, Kips Bay, Gramercy Park, \\
          &  Sutton, Tudor, Medical City, Stuy Town \\
    9     & Meat packing district, Greenwich Village, \\
          & West Village, Soho, Little Italy, China Town,\\ 
          & Civic Center, Noho  \\
    10    & Lower East Side, East Village, ABD Park, \\
          & Bowery, Two Bridges, Southern tip, White Hall,\\
          & Tribecca, Wall Street  \\
    \hline
    \end{tabular}%
  \label{tab:towns}%
\end{table}%

\begin{figure}[htbp]
	\centering
	\includegraphics[width=1.1\linewidth]{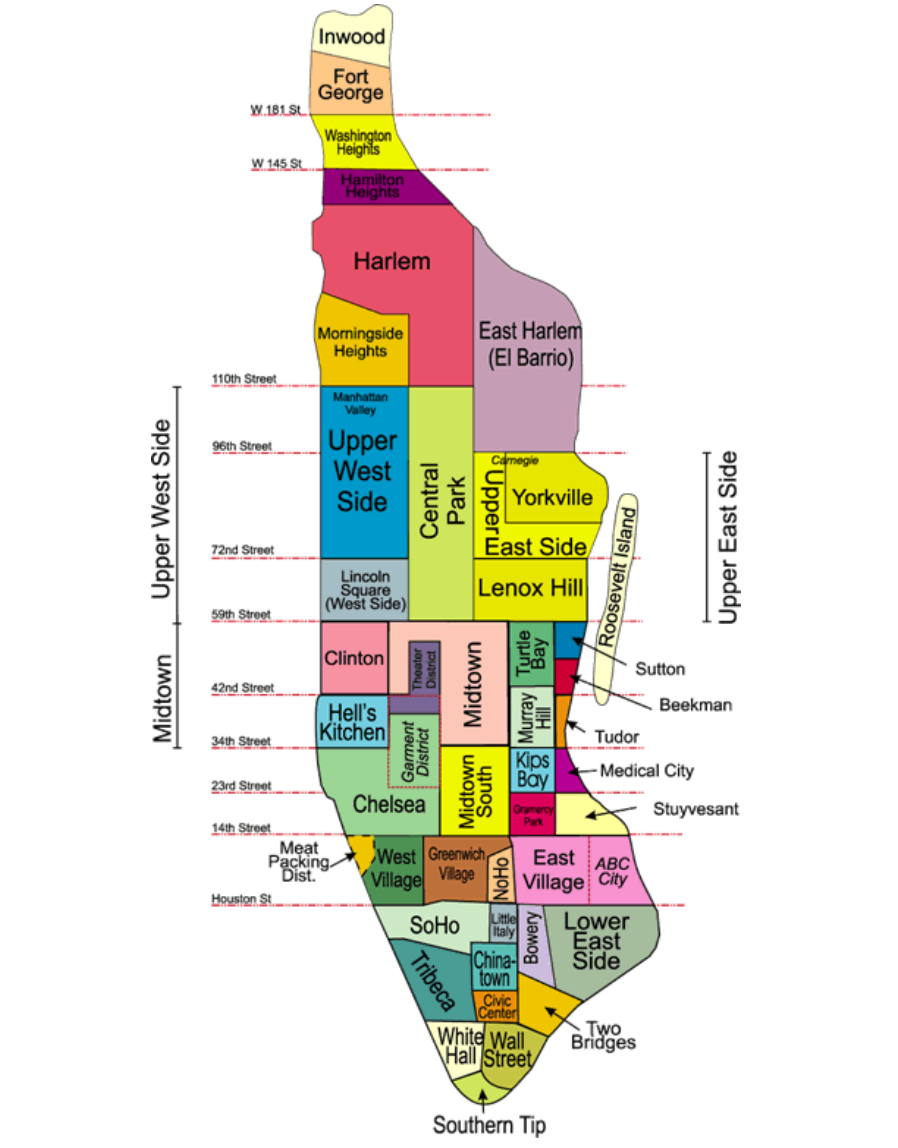}
	\caption{Towns in Manhattan, New York.}
	\label{fig:taxi}
\end{figure}

\end{document}